# A relic sketch extraction framework based on detail-aware hierarchical deep network

Jinye Peng, Jiaxin Wang, Jun Wang*, Erlei Zhang*, Qunxi Zhang, Yongqin Zhang, Xianlin Peng and Kai Yu

**Abstract**

As the first step of the restoration process of painted relics, sketch extraction plays an important role in cultural research. However, sketch extraction suffers from serious disease corrosion, which results in broken lines and noise. To overcome these problems, we propose a deep learning-based hierarchical sketch extraction framework for painted cultural relics. We design the sketch extraction process into two stages: coarse extraction and fine extraction. In the coarse extraction stage, we develop a novel detail-aware bi-directional cascade network that integrates flow-based difference-of-Gaussians (FDoG) edge detection and a bi-directional cascade network (BDCN) under a transfer learning framework. It not only uses the pre-trained strategy to extenuate the requirements of large datasets for deep network training but also guides the network to learn the detail characteristics by the prior knowledge from FDoG. For the fine extraction stage, we design a new multiscale U-Net (MSU-Net) to effectively remove disease noise and refine the sketch. Specifically, all the features extracted from multiple intermediate layers in the decoder of MSU-Net are fused for sketch predication. Experimental results showed that the proposed method outperforms the other seven state-of-the-art methods in terms of visual and quantitative metrics and can also deal with complex backgrounds.

## 1. Introduction

Painted cultural relics inherit the cultural essence of a country and provide valuable materials for historical research. Due to environmental changes and human damage, painted cultural relics are being destroyed in varying degrees around the world. Obtaining a drawn sketch is an important step in the protection and restoration of painted cultural relics. The sketch reflects the original ideas of the painters and the main structure and content of the painted images [1, 2, 3, 4]. Traditionally, the sketch is depicted manually by professional copyists, which is not only time-consuming but also affected by the different painting skills of different copyists [5, 6, 7]. To assist archaeologists and improve the accuracy of the sketch extraction, computer-aided sketch extraction for the painted images has been investigated [8, 9, 10].

Most of the existing computer-aided sketch extraction methods are based on edge detection techniques [8, 9, 11, 12, 13], which can be mainly divided into the three categories described below.

(1) Gradient-based algorithms [14, 15]. Typically, these algorithms compute the gradient information of the intensity or color to detect edges. Liu et al. [8] proposed an interactive sketch generation method to extract image outlines and learn the styles from examples. He et al. [9] generated the sketch by extracting hierarchical blocks and replaced the missing content with interactions. Kang et al. [14] proposed a flow-based anisotropic filtering framework (FDoG) to extract sketches. Sun et al. [10] designed an automatic generation sketch system based on the FDoG algorithm. Sun et al. [11] combined heuristic routing and high-frequency enhancement to collaboratively present a complete mural sketch. Xu et al. [12] used threshold segmentation and edge detection algorithms to combine gray information and edge information for sketch extraction. These methods can effectively extract landscape details in painted cultural relics. However, they are sensitive to areas with dense gradient changes. Due to the different degrees of disease, such as armor, chapped, faded [16, 17], and so on, painted cultural relic images always have complex backgrounds. With the above-mentioned methods, it is easy to extract noise in areas with serious disease. Moreover, the lines that do not have a significant gradient change will be lost, which makes the extracted sketches incoherent.

(2) Learning-based algorithms [18, 19, 20, 21, 22, 23]. These algorithms manually design features, such as intensity, gradients, and textures, to detect the edges through complex paradigm learning. Zitnick et al. [19] used the straight lines and T-junctions structures in local image patches to learn the edge detectors. Hussein et al. [24] proposed a hybrid optimization model combining particle swarm optimization with a local search algorithm for sketches. Qi et al. [25] proposed a learning ranking strategy that uses perceptual grouping to automatically generate sketches. These methods fuse prior knowledge into the sketch extraction and are robust to scattered lines and complex noise. However, they are developed based on the manual features, which mea-ns that they do not have sufficient differentiation to recognize sudden changes in complex relic images.

(3) CNN-based algorithms [26, 27, 28, 29, 30, 31, 32, 33]. These algorithms have shown promising edge detection performance with automatic features learning. Xie et al. [28] proposed an edge detection model with an overall nested structure. Liu et al. [34] obtained richer edge features by fusing convolutional intermediate layers. Hu et al. [31] added auxiliary branches to assist in extracting powerful advanced features for edge detection. He et al. [33] proposed a Bi-Directional Cascade Network (BDCN) and used scale enhancement modules to enrich edge features. Pan et al. [13] first proposed a CNN-based image fusion method for the sketch of the Dunhuang murals. These methods can automatically repair broken lines and suppress noise by deep feature learning. However, most of these deep learning approaches require large amounts of data to train the models and painted cultural relics data is limited. On the other hand, the CNN-based methods pay more attention to global characteristics while losing the detail information during deep propagation, which leads to blurred lines.

In this paper, we propose a detail-aware hierarchical neural network for accurate sketch extraction (as shown in Fig. 1). The proposed sketch extraction framework consists of two stages: coarse extraction and fine extraction. For the coarse extraction stage, we first designed a detail-aware BDCN model based on transfer learning and propose a novel fusion weighted loss function that incorporates the traditional edge detection algorithm (FDoG) and the BDCN. It not only keeps the advantage of traditional edge detection but also uses the edge information as the prior knowledge to guide the deep network focus on the extraction of detail features. Then, the results of coarse extraction are used as the input of the fine extraction stage. We designed a multi-scale U-Net (MSU-Net) model that fuses all the features extracted from multiple intermediate layers in the decoder to suppress disease and remove blur in the sketch. This framework combines the advantages of the traditional edge detection and deep learning methods as well as achieves promising sketch extraction of the painted cultural relics.

Our main contributions can be summarized as follows:

1. We are the first to develop a hierarchical deep network framework for the sketch extraction of the painted cultural relics. Different from the previous method, we designed a hierarchical structure to perform coarse sketch extraction and fine sketch extraction. The proposed hierarchical deep network framework is promising in its ability to extract a clear, coherent, and complete sketch of painted cultural relics.
2. We designed a novel detail-aware BDCN model for coarse sketch extraction, which combines the advantage of traditional edge detection in detail extraction and deep feature learning in edge detection and noise suppression. It not only solves the problem of limited relics data, but also produces coarse sketches with good coherence and complete information for further fine sketch extraction.
3. We designed a novel MSU-Net combining multiscale high-level features for the fine sketch extraction. Based on the results of coarse extraction, MSU-Net fused different levels of feature in the deep network, which not only suppresses complex diseases but also refines the lines and removes the blur.

The remainder of this paper is organized as follows: Section 2 introduces related edge detection work. Section 3 describes the proposed methods in this paper. All the experimental results and analyses are shown in Section 4. A summary and some closing remarks are made in Section 5.

## 2. Related Works

### 2.1 Flow-Based Difference-of-Gaussians (FDoG)

FDoG [14] is a traditional edge detection algorithm. It uses a kernel-based nonlinear smoothing of vector field meth-od to construct the edge flow field to highlight the edge tangential direction. Then, a linear DoG filtering [35] is applied in the gradient direction while moving along the edge flow. The last step accumulates the filtered responses along the flow to extract the lines. The edge tangential flow construction filter definition is given in Eq. 1.

$$\mathrm{t}^{\mathrm{new}}(x) = \frac{1}{k} \sum_{y \in \Omega(x)} \phi(x,y) \mathrm{t}^{\mathrm{cur}}(y) \omega_s(x,y) \omega_m(x,y) \omega_d(x,y) \tag{1}$$

where $\Omega(x)$ denotes the neighborhood of x; $\mathrm{t}^{\mathrm{cur}}(y)$ denotes the normalized tangent vector at y; $\omega_s$ is the spatial weight function, given the neighborhood size; $\omega_m$ is the magnitude weight function, giving the greater weight for higher gradients to ensure the main edge direction; and $\omega_d$ is the direction weight function, which measures the gradient difference between the neighborhood pixels, and uses the symbol function $\phi(x, y) \in \{-1, 1\}$ to make a close alignment of the vectors.

Compared with other traditional edge detection algorithms, the FDoG algorithm preserves the significant edges and guides the weak edges to follow the direction of the significant edges in the neighborhood. It can maintain the line coherence as much as possible and extract the details completely. However, the FDOG algorithm is based on gradient information. Therefore, it may produce dense noise lines in complex background regions with the disease. To improve line coherence, it is inevitable to smooth the sketch extracted by the local core, which may cause abstraction and distortion. Overall, FDoG can extract complete details while it easily suffers from complex noise and partial abstraction.

## 2.2 Bi-Directional Cascade Network (BDCN)

As shown in Fig. 1, the structure of the BDCN [33] for edge detection is based on VGG16 [36] without three fully connected layers and the last pooled layer. It consists of five convolution layers stages. Each stage is called an Incremental Detection Block (ID Block) which gradually expands the receptive field with the pooling layer to get different scales. Then, to predict different scales of edges, it implements layer-specific supervision through a bi-directional cascade structure.

Each ID blocks is formed by inserting a Scale Enhancement Module (SEM) [37] into several convolution layers corresponding to VGG16. Where SEM consists of several dilated convolutions of different dilation rates. Then the SEM output in each ID block is fused and fed into two $1\times1\times21$ and $1\times1\times1$ convolutions respectively to generate two edge predictions $P^{s2d}$ and $P^{d2s}$ at this scale. For the two complementary supervised learning at the scale of s, it is defined by Eq. (2) to achieve a specific layer to be learned by specific supervision. Where $P^{s2d}$ is the prediction from the shallow to the deep, and $P^{d2s}$ is the prediction from the deep to the shallow. Then upsampling the deep feature maps to the original image size to generate edge predictions for the corresponding scale. Finally, use a $1\times1\times1$ convolutional layer to fuse the multi-scale edge prediction to generate the final edge prediction.

$$\begin{aligned} Y_s^{s2d} &= Y - \sum_{i<s} P_i^{s2d}, \\ Y_s^{d2s} &= Y - \sum_{i>s} P_i^{d2s}, \end{aligned} \tag{2}$$

The BDCN method can learn the multiscale feature through dilated convolution [37] and it has advantages in the edge detection task because of layer-specific learning. However, deep BDCN requires large amounts of data to train the models, which is a challenge for the limited painted cultural relics data. Furthermore, it may appear blurry and there may be a loss of details at the edge as the depth of the network becomes deeper.

### 2.3 U-Net

The U-Net network first appeared in the field of medical image segmentation [38], which is an end-to-end network based on the fully convolutional network. The structure of U-Net is composed of a symmetrical encoder and decoder. The encoder uses convolution and pooling layers to mine deep image features. The decoder combines the upsampling with the feature map of the encoder pooling layer, and then upsampling to the size of the original picture layer by layer. Especially, the high-resolution position feature of the encoder is combined with the high-level abstract feature of the decoder via skip connection. The U-Net network has been widely used for image denoising [39, 40] and segmentation [41, 42, 43], and has shown promise in this field.

## 3. Method

The sketch extraction framework proposed in this paper is shown in Fig. 1, which consists of coarse extraction and fine extraction. In the coarse extraction stage, a detail-aware BDCN method is developed that uses the traditional edge detection method as prior knowledge and pre-trains the model with the natural image dataset. Then, the detail-aware BDCN is fine-tuned by the painted cultural relics images. In the fine extraction stage, the outputs of the detail-aware BDCN are used as the inputs of the proposed MSU-Net, which fuses multiscale features for refining the sketch and denoising.

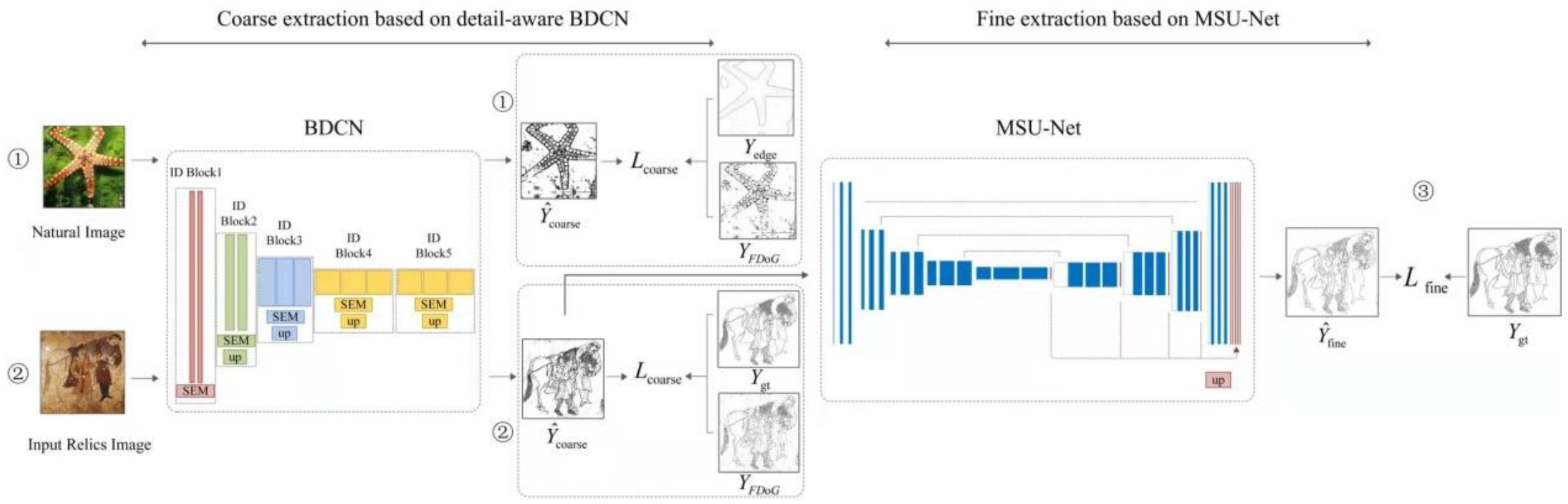

Figure 1: The proposed sketch extraction framework.

### 3.1 Coarse Extraction Based on Detail-Aware BDCN

In Fig. 2, there are four natural images along with their true edge labels of the BDCN and sketches extracted by the FDoG algorithm. It can be seen that the typical BDCN meth-od focuses on the contour extraction of the object in the images, while it loses plenty of image details. As the network deepens, the features learned by BDCN will become more abstract. FDoG algorithm extracts the edge in the image based on using the gradient, which is sensitive to differences of color and texture. The sketch from the FDoG algorithm contains rich details while it suffers from noise and fake edges. Thus, this paper proposed a novel detail-aware BDCN integrating existing

BDCN and FDoG algorithms to complement each other. It firstly used natural images and extractions by FDoG to pre-training the model, which not only solved the problem of the limited relics data, but also utilized low-level features of natural images including shapes, texture, etc. Then, we fine-tuned the deep network on relics data with the extractions by FDoG, which not only used the prior knowledge to help the deep network focus on the learning of boundaries and details but also captured deep abstract features.

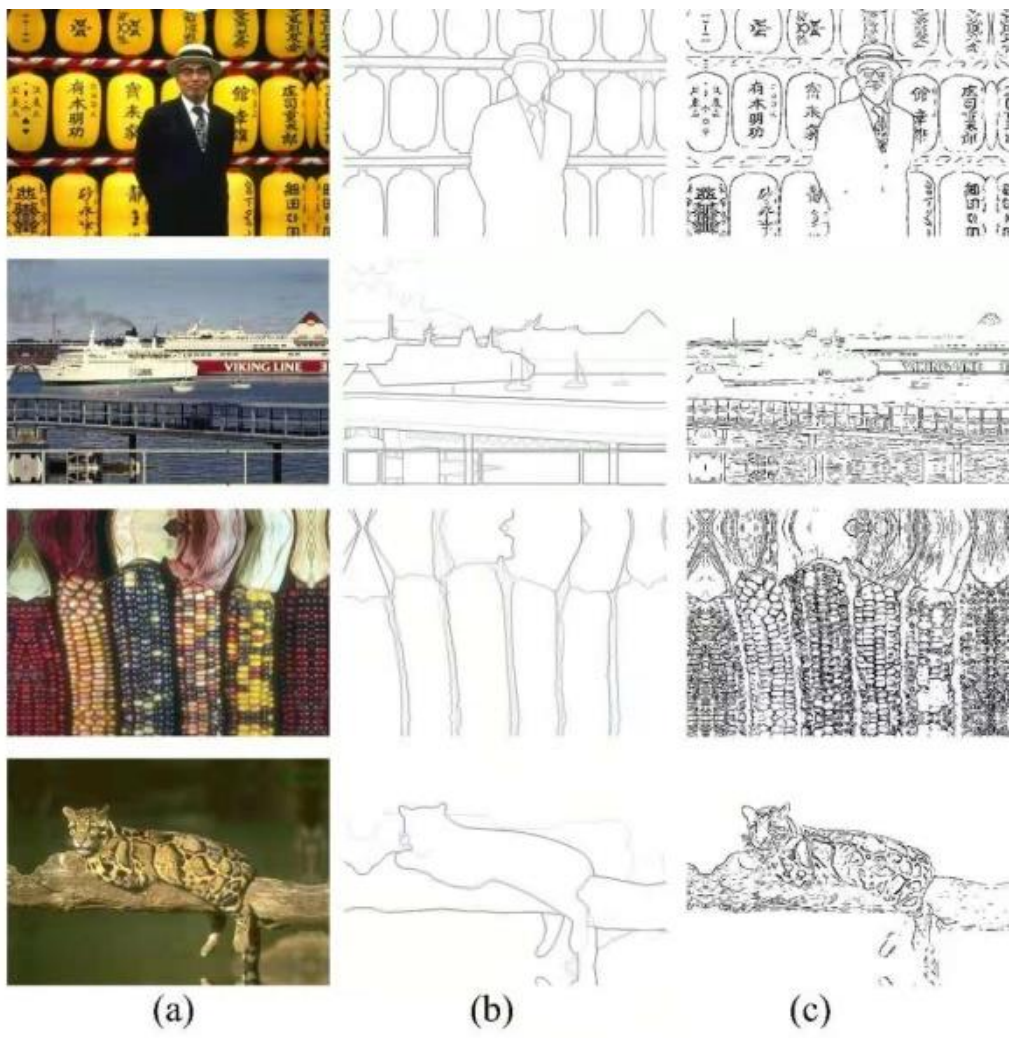

Figure 2: Edge detection on natural images. (a) Original natural images; (b) True edge labels of BDCN; (c) Sketches extracted by FDoG algorithm.

We design a learning method that uses traditional FDoG edge detection algorithms to assist in the enhancement of detail awareness, which can force deep learning to pay attention to details to extract sketch with more complete information. Firstly, the FDoG algorithm is used to extract sketch, and it is combined with the labels of the edge detection network to form a supervisory pair. And we designed a weighted loss function to constrain the learning of the network:

$$L_{coarse} = \alpha L(P_{coarse}, Y_{FDoG}) + \beta L(P_{coarse}, Y_{edge}) \quad (3)$$

Where $Y_{FDoG}$ denotes the sketch extracted by the FDoG algorithm which contains rich details, $Y_{edge}$ denotes the label of the edge detection, $P_{coarse}$ denotes the prediction of coarse extraction of the sketch. $\alpha$ and $\beta$ are the weights of the predicted sketch between the FDoG extraction sketch and the label of the edge detection in the loss calculation, respectively.

In this study, a prediction was given for each pixel. For the ground truth, the pixels on the sketch were defined as the positive samples, that is, $Y_{+} = \{y_j, y_j > \gamma\}$. The non-sketch pixels were defined as the negative samples, that is, $Y_{-} = \{y_j, y_j = 0\}$. Besides, $\gamma$ was the threshold for dividing pixels into positive and negative samples. However, the distribution of sketch and non-sketch pixels was heavily biased. Lin et al. [44] proposed a focal loss and designed two balance factors to solve the problems of hard and easy sample imbalance and positive and negative sample imbalance. Since our positive and negative pixels were seriously imbalanced, we adopted the method of adding a class balance factor in the focal loss, that is, using the following class-balanced cross-entropy loss function to define Eq. 3:

$$L(\hat{Y},Y) = -\mu \sum_{j \in Y_-} \log(1-\hat{y}_j) - \nu \sum_{j \in Y_+} \log(\hat{y}_j) \tag{4}$$

Where $\hat{Y}$ is the predicted sketch and $\mu = |Y_+|/(|Y_+|+|Y_-|)$, $\nu = |Y_-|/(|Y_+|+|Y_-|)$ is used to balance the positive and negative samples.

During the training stage, we used the natural image dataset to pre-train the detail-aware BDCN and the relic image dataset to fine-tune the model. In fine-tuning, we replaced $Y_{\text{edge}}$ with the ground truth of painted cultural relics for training, which provided more relics details and also made the sketch extraction model more in the style of the cultural relics.

Coarse extraction based on detail-aware BDCN combines the advantages of deep feature learning and traditional edge detection algorithms, which overcomes the problem of detail loss and produces a coarse sketch. Because FDoG easily suffers from noise, it is more likely to introduce noise into the prior detail knowledge $Y_{\text{FDoG}}$, which leads to fake lines in the coarse sketch. Moreover, there are various scales of the object in the painted cultural relics images. According to our experience, detail-aware BDCN generates blur and artifacts in some pixels because of the network directly upsampling all the features to the same scale for prediction. To overcome these problems, we propose an MSU-Net method to further refine the coarse sketch.

### 3.2 Fine Extraction Based on MSU-Net

The typical U-Net follows the encoder-decoder paradigm for feature learning, which is a promising image denoising and reconstruction method [38, 39, 45, 40, 46]. In this study, we considered the disease suppression and sketch extraction tasks in the painted cultural relics to be a denoising problem and reconstruction tasks. Instead of using the typical U-Net for image reconstruction directly, we propose an MSU-Net to refine the coarse sketch. Typical U-Net commonly uses feature maps from the last layer in the decoder path to make the prediction. However, the features from convolutional learning will gradually become abstract as the network became deeper. Many meaningful multiscale detail features may be lost [34, 47, 48, 49], which should be considered in the sketch extraction.

The structure of the proposed MSU-Net is shown in Fig. 3. The MSU-Net keeps the main structure of the typical U-Net in which the features of the encoder path are connected into the decoder path for the fusion of features from the encoder path and decoder path. The fused features can contain the different size receptive fields information from the encoder path. In the decoder path, different from like U-Net directly upsampling the fused feature, we add a multiscale feature fusion path. In this path, the fused features are copied and connected to a $1 \times 1 \times 1$ convolutional layer. Then, a sketch prediction is generated for each scale (called side-outputs, as shown in Fig. 3) by using a transposed convolutional operator to upsample all the feature maps from the convolutional layer to the same size of the original image. The side-outputs directly access the prediction results of different scales. Due to the different receptive fields of the convolutional layer of the decoding path, when the side-output predicts from shallow to deep, it focuses on small details and gradually captures larger targets, which can directly predict both low-level information and object-level information. Finally, in order to fuse the results of different levels of prediction, the output feature map of the decoder path and all the side-outputs are combined and fed into a 1 x 1 x 1 convolutional layer for sketch prediction. Based on this, we enhanced the representation of multi-scale features, which is beneficial to our capture of multi-scale disease and the learning of edge and details. Note that the 3×3 convolutional layers used to extract features in MSU-Net are set to padding = 1 to ensure that the output is the same size as the input image.

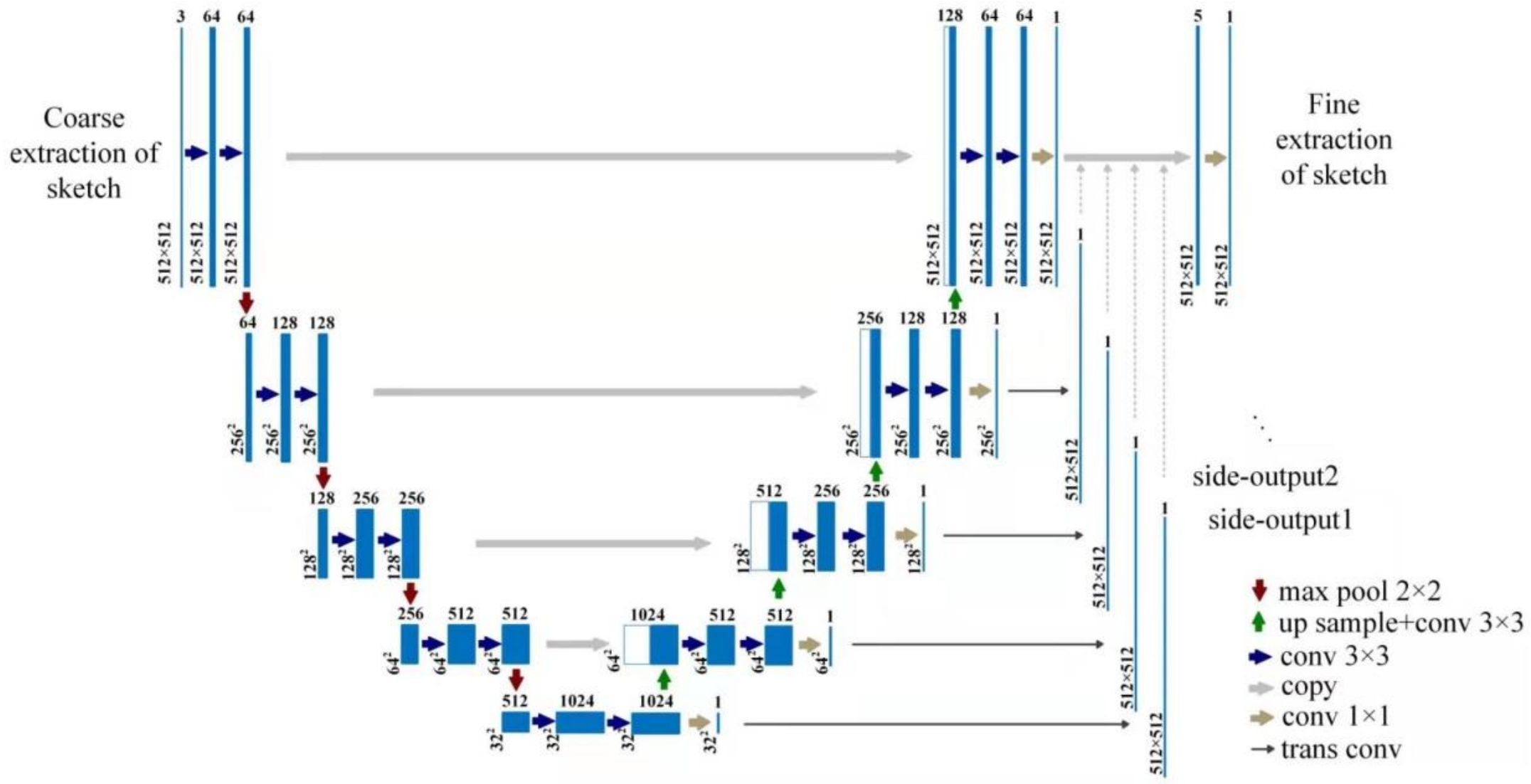

Figure 3: The detailed architecture of MSU-Net.

The highlights of MSU-Net are that it not only utilizes feature maps from multiscale receptive fields in the encoder path to suppress the effect of disease in the relic image but it also utilizes multiscale feature maps in the decoder path for de-blurring and refining the sketch.

### 3.3 A Hierarchical Deep Network Framework for Sketch Extraction

The hierarchical deep network framework concatenates the detail-aware BDCN and MSU-Net, as shown in Fig. 1, in which there are three steps during the training stage: 1) The detail-aware BDCN model is pre-trained with the natural image dataset. Since painted cultural relics data are limited, it is difficult to directly train a deep network with good performance. We used a natural image dataset with the truth edges and the extracted sketches by the FDoG algorithm to pre-train a detail-aware BDCN model as described in Section 3.1. 2) The painted cultural relic images were used to fine-tune the detail-aware BDCN model as the coarse extraction model. In this step, we first used the FDoG algorithm to extract the edge from the painted cultural relic images and then combined it with the true sketch delineated by the expert to fine-tune the detail-aware BDCN model. 3) We trained the MSU-Net as the fine sketch extraction as described in Section 3.2. The coarse sketches from the detail-aware BDCN were used as the inputs of MSU-Net and the true sketches were used as the outputs of MSU-Net. MSU-Net was trained with the proposed fusion loss function in Eq. 6. We finished the training of the whole framework with the above three steps. At the test stage, test samples were passed through the two stages in the framework for sketch extraction.

In the proposed framework, transfer learning was used to solve the limited training dataset problem and the FDoG and BDCN were integrated for coarse sketch extraction, which preserves both contour and details at the same time. Furthermore, MSU-Net fused multiple predictions with different scales for the refinement of relic sketches.

## 4. Experimental Results and Discussion

### 4.1 Dataset

**Natural image dataset:** We used the public natural image dataset, BSDS500 [18], to pre-train the detail-aware BDCN model. The BSDS500 dataset contains 500 images and reference edge labels.

**The painted cultural relics images:** We collected 53 relic images from Fengguo Temple, Qianling, and the Dunhuang murals, which contain complex scenes like Buddha images and Tang Tomb. All the reference sketches of the relic images were delineated by the experienced experts from the Shaanxi History Museum. The sizes of the images range from $160\times160$ to $1000\times1000$. In our experiments, we selected 41 images and cropped them into 82 sub-images (to increase the number of images) for training the model. During training, we adopted the data augmentation strategy [28, 50, 30] to make our model more robust. To validate the performance of the proposed method, the remaining 12 images were used for testing, which contained two scenes: clean background images (e.g., Echographic of Dunhuang murals, as shown in the first and second rows of Fig. 4) that have less disease and complex background images (e.g., "Polo Painting" and "Preparing horses painting," as shown in the third and fourth rows of Fig. 4) that have cracks and shedding.

**Hyperspectral painted cultural relics images:** To validate the generalization ability of the proposed method, we extended an experiment, passing a characteristic band of the hyperspectral image through two stages in the hierarchical sketch extraction framework for testing. We used hyperspectral imaging equipment, 'SOC710,' to collect more than 20 relic images from Fengguo Temple and Qianling. Based on different data characteristics, we manually selected different bands with more obvious sketch information as the test data, for example, "Qianling hunting trip painting," as shown in Fig. 8(a), and its 755th band as shown in Fig. 8(a1) ; "Buddha’s seat of lotus flower painting," as shown in Fig. 8(b), and its 740th band as shown in Fig. 8(b1); "The Eighteen Arhats in the Fengguo Temple," as shown in Fig. 8(c) and Fig. 8(d), and their 6th band after Minimum Noise Fraction Rotation (MNF) [51], as shown in Fig. 8(c1) and Fig. 8(d1).

### 4.2 Experimental Setups

To demonstrate the effectiveness of the proposed method, seven state-of-the-art algorithms, of which there were six edge detection algorithms, including Canny [52], FDoG [14], Edge-Boxes [19], HED [28], RCF [34], BDCN [33], and the related algorithm U-Net [38], were chosen for comparison.

The root mean squared error (RMSE) [53], structural similarity index (SSIM) [53], and average precision (AP) [15] were used to evaluate the results of the sketch extraction, which respectively measure the error, structural similarity, and average accuracy between the extracted sketch and the ground truth. The larger the value of SSIM and AP are, the better the performance. The smaller the value of RMSE is, the better the performance.

Furthermore, we investigated and analyzed the effect of each module on the performance of the proposed method, including detail-aware BDCN, MSU-Net, and several key parameters. Finally, an experiment on hyperspectral images was conducted to show the generalization ability of the proposed method.

## 5. Experimental Results

### 5.1 Comparison with Other Works

The results of our method and the compared methods are given in Tab. 1. For the HED, RCF, and BDCN methods, the results of two cases are reported 1) without "*", which means that we use the methods trained on natural images and tested on our testing dataset, and 2) with "*", which means that we trained the methods on our training dataset and tested them on our testing dataset.

Table 1: Comparison of existing edge detection works and related works.

| Method | RMSE | SSIM | AP |
|---|---|---|---|
| Canny | 0.3615 | 0.8268 | 0.4172 |
| FDoG | 0.2561 | 0.8993 | 0.5491 |
| Edge-Boxes | 0.3747 | 0.6367 | 0.3569 |
| HED | 0.3237 | 0.8105 | 0.3432 |
| RCF | 0.3206 | 0.7598 | 0.3561 |
| BDCN | 0.3186 | 0.8263 | 0.3887 |
| HED* | 0.2938 | 0.7531 | 0.5321 |
| RCF* | 0.2969 | 0.7216 | 0.5245 |
| BDCN* | 0.2436 | 0.8491 | 0.5541 |
| U-Net* | 0.2280 | 0.9931 | 0.6678 |
| **Ours** | **0.1956** | **0.9963** | **0.7846** |

In terms of the RMSE and AP metrics, deep learning-based methods (HED*, RCF*, BDCN*, and our method) outperformed some traditional edge detection methods (Canny) and the learning-based method (Edge-Boxes) on our cultural relics data. In particular, U-Net* and our method outperformed the compared traditional methods (Canny, FDoG) and learning-based method (Edge-Boxes) in three metrics, which proves the advantage of deep learning-based methods on sketch extraction. Second, the proposed method outperformed the other six edge detection methods in three performance metrics, which were around 5~17%, 1~35%, and 23~44% in terms of RMSE, SSIM, and AP, respectively. Specifically, the proposed method achieved AP values around 23% higher than those of the BDCN* and FDoG algorithms, which indicates that the proposed method had a visible improvement in the completeness and accuracy of sketch extraction compared to the other two methods. The proposed method outperformed the FDoG and BDCN* algorithms by about 6% in RMSE, which indicates that the extracted sketches by our method had a smaller average error in the whole images. Moreover, the proposed method achieved the highest value in the SSIM, which means that the extracted sketches had the best structural similarity with the ground truth.

In addition, U-Net* outperformed BDCN* by about 14% in SSIM, which indicates that the sketches extracted by U-Net had a higher structural similarity with the relic style, showing the capacity of U-Net to refine sketches. The proposed method reached the highest in SSIM, and it was superior to U-Net* by about 11% in AP value and about 3% in RMSE value. Compared with U-Net*, which only uses relic data for training, our hierarchical framework uses a coarse extraction model trained on a large number of natural data and then refines it, which effectively solves the problem of insufficient data to achieve better performance.

The extracted sketches and some partially enlarged details of the four images by the different methods for the visual evaluation are shown in Fig. 4. As shown, FDoG suffered from noise and generated discrete points and broken lines, especially in the third and fourth images, which have serious diseases. The extraction results of the Edge-Boxes are blurred and many important details were lost in the four images. The sketches from RCF suppressed the effect of complex noise; however, they appeared blurry in the four images and many details were lost, especially the facial features in the first two images and the boots texture in the fourth image. The sketches from BDCN* that were trained with relic data are relatively complete, but areas with denser details were easily blended and blurry. The extraction results of U-Net* show insufficient learning, which not only resulted in the loss of much information but also caused them to suffer from noise. Compared with those results, the sketches from the proposed method are much clearer; they are not only complete and continuous but also have vivid details. These results illustrate the advantages of our method on disease suppression and detail extraction for relic sketch extraction.

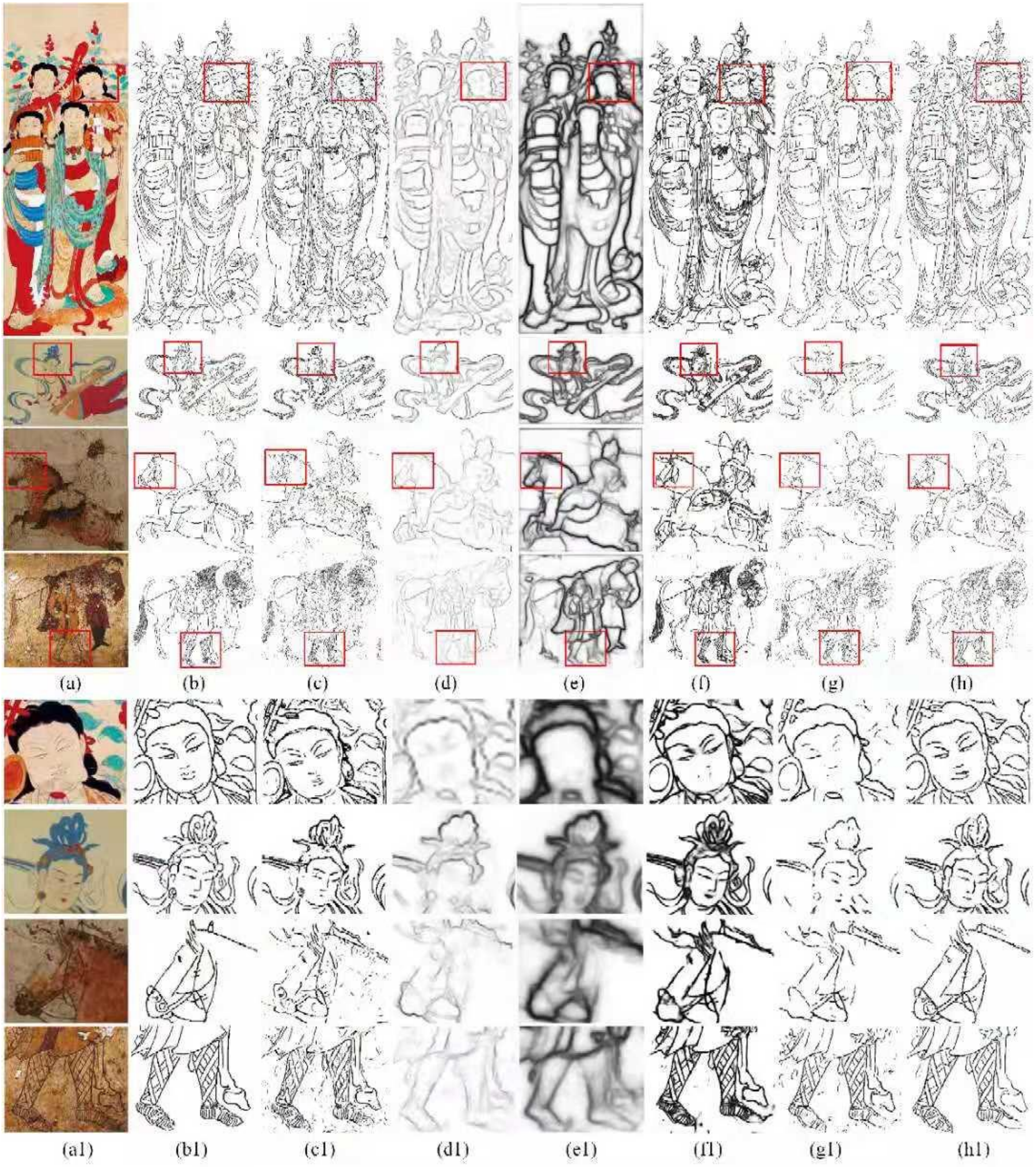

Figure 4: (a) The original painted cultural relics image; (b) Ground truth; (c) FDoG; (d) BDCN; (e) Our coarse extraction. The right figures (a1-e1) are the corresponding partially enlarged details.

Overall, the proposed method achieves much better performance for relic sketch than the previous methods in the objective and visual evaluation.

## 5.2 Analysis of Each Module

Two sets of experiments were designed to verify the effect of the detail-aware BDCN in coarse extraction and MSU-Net in fine extraction. Furthermore, we discuss the weight parameters of FDoG in the coarse extraction and the performance of fuse side-output layers in MSU-Net.

### 5.2.1 The Effect of Detail-Aware BDCN

Detail-aware BDCN was designed on the deep edge detection algorithm (BDCN)

and guided by the gradient-based edge detection algorithm (FDoG) to focus on detail feature learning. Specifically, we designed a weighted loss function to guide the network learning during the training stage, where $\alpha$ and $\beta$ were used to balance the loss between the prediction and the edges obtained by FDoG and the loss between the prediction and the true edge. Sketches from the coarse extraction stage were compared with different parameters in the weighted loss function to verify the effect of detail-aware BDCN, as shown in Tab. 2 and Fig. 5.

Table 2: The performance of weight parameter for coarse extraction sketch.

| $\alpha/\beta$ | 0/1 | 0.05/0.95 | 0.1/0.9 | 0.2/0.8 |
|---|---|---|---|---|
| Recall | 0.459 | 0.731 | 0.916 | 0.918 |
| RMSE | 0.337 | 0.320 | 0.341 | 0.374 |

In this experiment, the weight ($\alpha$ and $\beta$) was varied to tune the effect of the FDoG algorithm on the detail-aware BDCN and recall [15] and RMSE were used as the performance metrics. Because recall is the number of correct pixels divided by the number of results that should have been extracted, the larger the value of recall is, the more correct pixels there are in the sketch. The smaller the value of RMSE is, the less noise in the sketch.

In Tab. 2, 0/1 denote the results from BDCN without the guidance of the FDoG algorithm. As shown, the recall value improved as the value of $\alpha$ increased. Specifically, there were obvious improvements in recall value when $\alpha$ increased from 0 to 0.05, which means that the deep network captures more details with the guidance of the FDoG algorithm. Increasing $\alpha$ from 0.1 to 0.2 yielded only minor improvements of the recall value, while the RMSE value rose. The reason for this could be that the FDoG algorithm is sensitive to noise in the original image, which may introduce noise in the results when FDoG has more effects. Thus, we set $\alpha$ =0.1 and $\beta$ =0.9 in our experiments for a balance of details extraction and noise suppression.

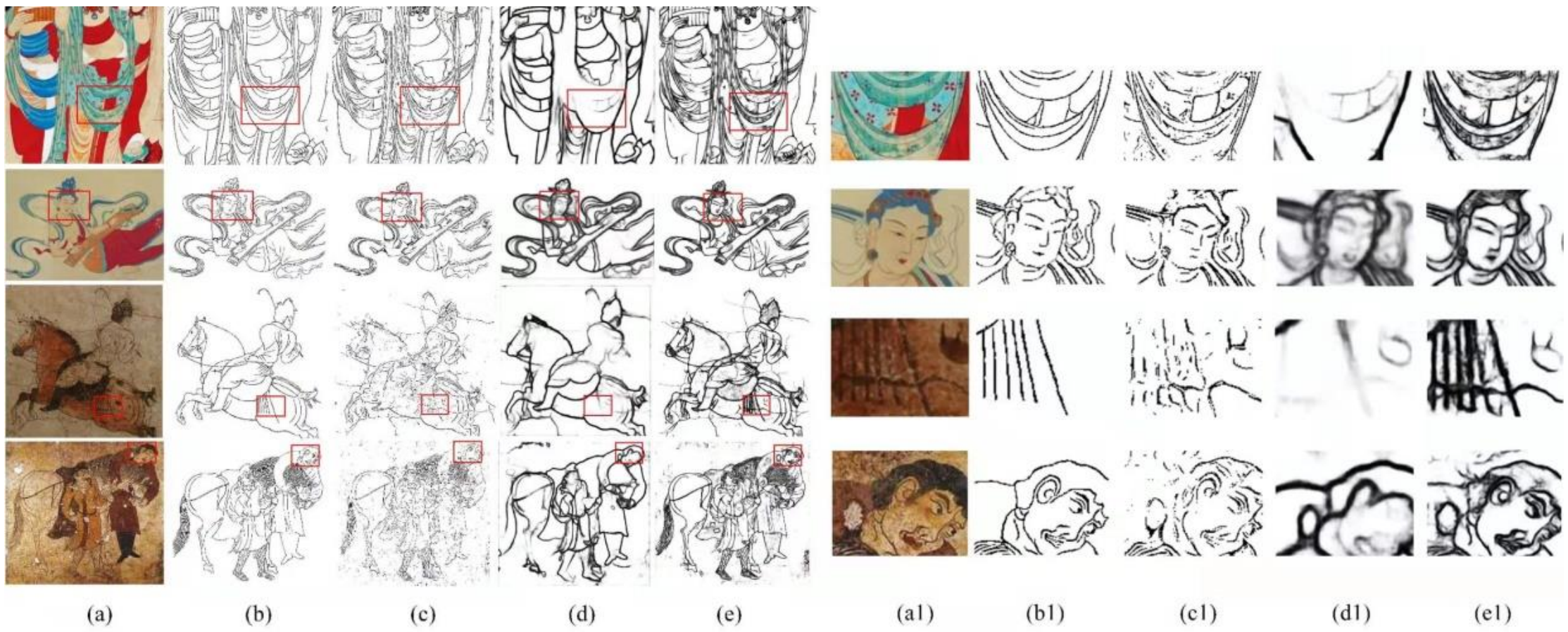

Figure 5: (a) The original painted cultural relics image; (b) Ground truth; (c) FDoG; (d) BDCN; (e) Our coarse extraction. The right figures (a1-e1) are the corresponding partially enlarged details.

Similar conclusions can be made from a visual evaluation, as shown in Fig. 5.

The results from the FDoG algorithm have lots of discrete points and broken lines because of noise and disease in the original images. Compared with the FDoG algorithm, the results from BDCN extract the contour of the object with continuous lines and have fewer effects from the noise and disease. However, the BDCN method lost many details in regions of interest and generated blurring lines in most of the edges. Compared with those two methods, the proposed detail-aware BDCN overcame those problems and generated a coarse sketch with rich details and clear outlines. Although the results from detail-aware BDCN have some blurs in local regions caused by the disease, it successfully utilized the advantages of the FDoG and BDCN algorithms to generate a coarse sketch for the next refinement.

### 5.2.2 The Effect of MSU-Net

To verify the effect of the proposed MSU-Net in the fine extraction, two group experiments were designed based on the coarse sketch extracted by detail-aware BDCN: (1) we used BDCN, U-Net and MSU-Net respectively as the fine extraction network to compare and verify the effect of MSU-Net to the sketch refinement and (2) we varied the structure of MSU-Net with different fusion strategy.

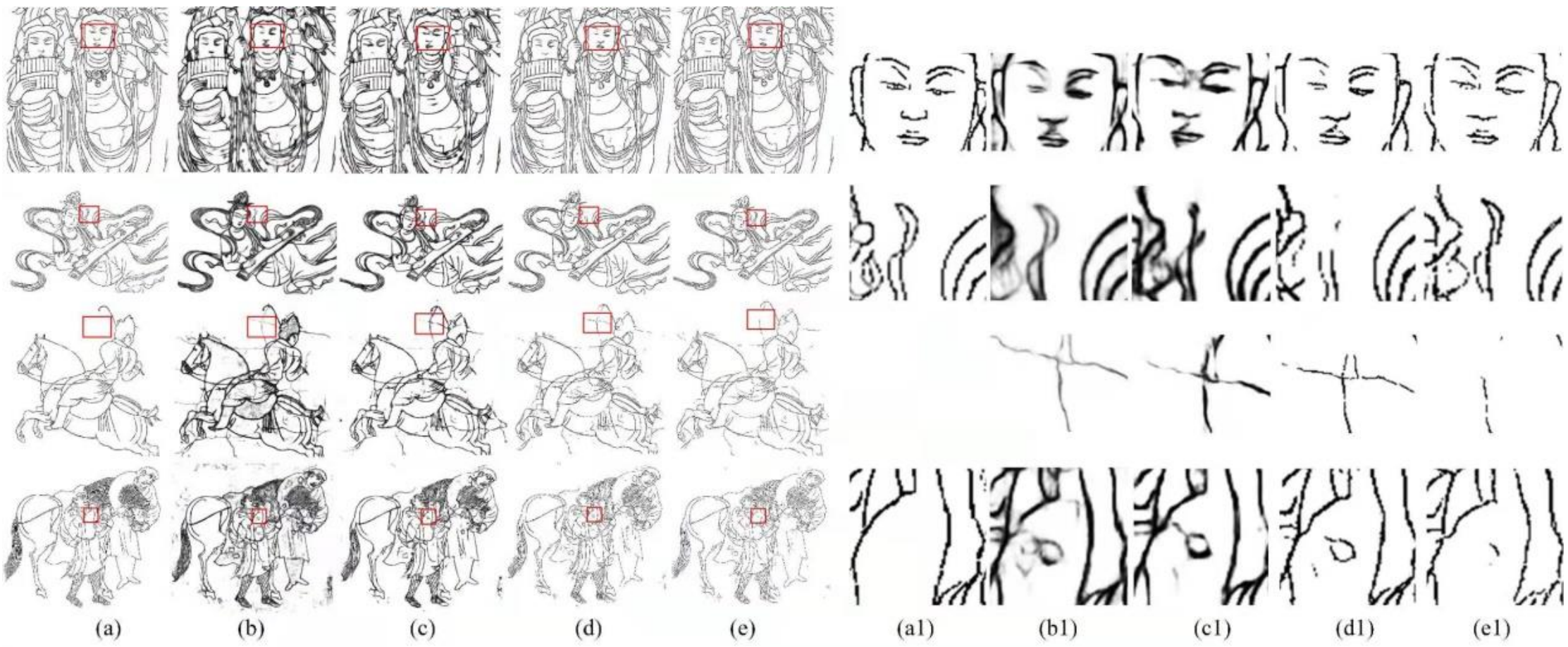

Figure 6: (a) Ground truth; (b) Coarse sketch; (c) Fine extraction by BDCN; (d) Fine extraction by U-Net; (e) Fine extraction by MSU-Net. The right figures (a1-e1) are the corresponding partially enlarged details.

The results produced by different fine extraction methods are shown in Fig. 6. The coarse sketches generated by the detail-aware BDCN have rich details of interest and outlines while containing blurred and fake lines caused by noise and disease. The refined sketches by BDCN still have blurred lines, especially in the densely detailed areas like the first and second images. U-Net and our MSU-Net had better performance than that of BDCN on the de-blurring, leading to more clear sketches in all the cases. Furthermore, we compared the effect of U-Net and MSU-Net. To verify the effect of refinement, we enlarged the details from the first and second images. MSU-Net had a better reconstruction effect on the sketches and retained a more complete sketch. Moreover, to verify the effect of disease suppression, we enlarged the cracks and shed disease areas in the third and fourth images. As shown, MSU-Net was less affected by noise and disease and produced fewer false lines in the sketches.

The reason for this could be that the proposed MSU-Net fuses the predictions from multiple scales, which overcomes the effects of the noise and disease. Overall, this experiment proved that the proposed MSU-Net could make up for the shortcomings of the detail-aware BDCN and generate satisfactory results through hierarchical sketch extraction.

Furthermore, we investigated the effect of MSU-Net structure by varying the fusion of the side-outputs, as shown in Tab. 3. Side-output5 means that only the last layer in the decoder path was used for prediction, that is, the same as traditional U-Net. Side-output5, 4 indicates that layer 5 and layer 4 were used in the decoder path for predictions, respectively, and then these two predictions were fused for the final sketch. The rest could be completed in the same manner.

Table 3: The performance of fuse side-output layers in MSU-net.

| fuse side-output layers | RMSE | SSIM | AP |
|---|---|---|---|
| Side-output5 | 0.2185 | 0.9921 | 0.6745 |
| Side-output5,4 | 0.2095 | 0.9897 | 0.6980 |
| Side-output5,4,3 | 0.2061 | 0.9788 | 0.7335 |
| Side-output5,4,3,2 | 0.2003 | 0.9905 | 0.7385 |
| **Side-output5,4,3,2,1** | **0.1956** | **0.9963** | **0.7846** |

From three evaluation metrics, it was found that the performance of MSU-Net generally improved as the number of the fused side-outputs increased. It achieved the best results when fusing all the side-outputs. We also show each side-output in Fig. 7, where there are different predictions with the variety of network's depth in the decoder path. These predictions focus on different receptive fields. The proposed MSU-Net effectively fused those predictions to make a better prediction that contains vivid details without being affected by noise and disease.

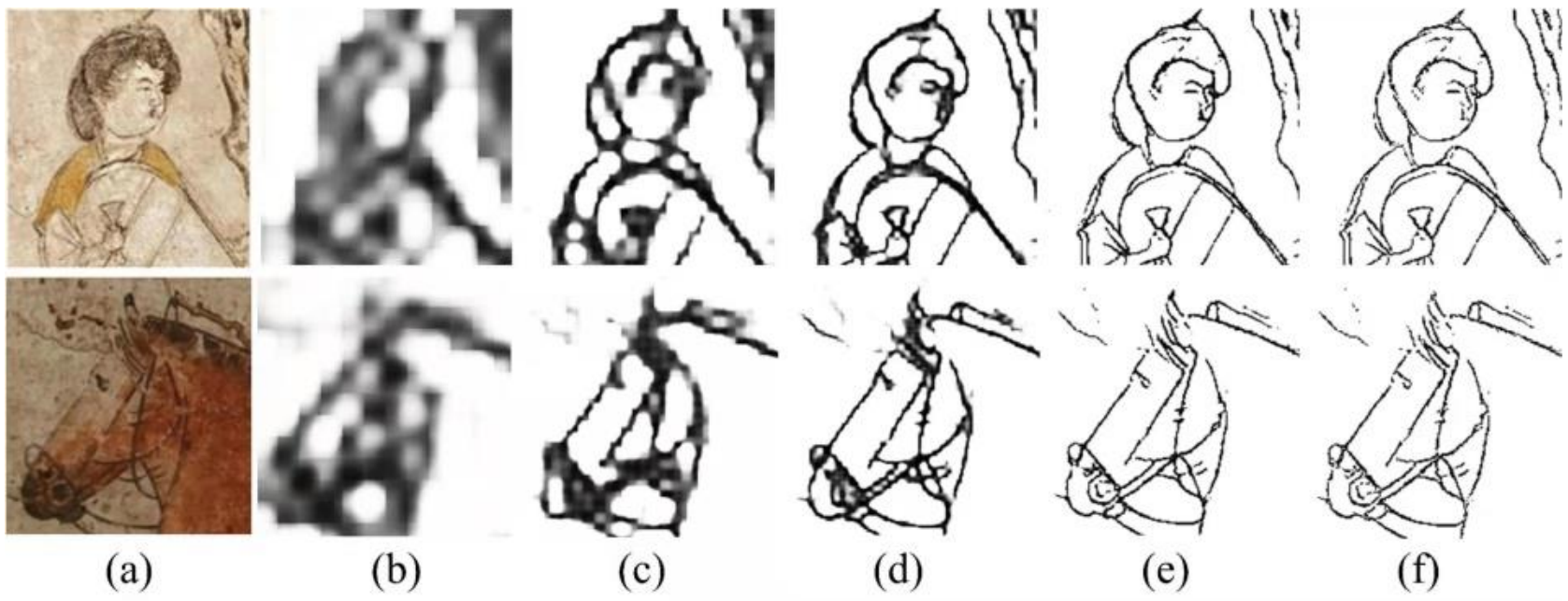

Figure 7: Several examples of the side-outputs of MSU-Net. (a) The original painted cultural relics image; (b) Side-output1; (c) Side-output2; (d) Side-output3; (e) Side-output4; (f) Side-output5. The first line and the second line are the "Lady painting" and the "Polo painting" in the tomb of Tang Wei imperial concubine, respectively.

### 5.2.3 Generalization Ability of the Proposed Method

Hyperspectral imaging technology can obtain the spatial and spectral information of each pixel at the same time. For painted cultural relics, hyperspectral images have advantages on the distinguish of substances with spectral-spatial information, which will greatly improve the ability to identify areas covered by dust, pollutants, and mine sketch information hidden under visible light. Due to the particularity of cultural relic images, hyperspectral imaging has been widely used in cultural relic research in recent years [54, 55, 56, 57, 58]. Therefore, we conducted experiments on hyperspectral images, combining the ability of hyperspectral technology to mine features to further validate the generalization ability of our method.

Hyperspectral images have a large number of bands and high information redundancy. For some noisy and complex images, we first performed MNF to reduce noise and then manually selected a band with obvious sketch information for testing. The selected band (Fig. 8(a1-d1)) could provide richer hidden information than visible light (Fig. 8(a-d)). Moreover, we passed it through two stages in the hierarchical sketch extraction framework for testing, which could make full use of the information to extract accurate sketches.

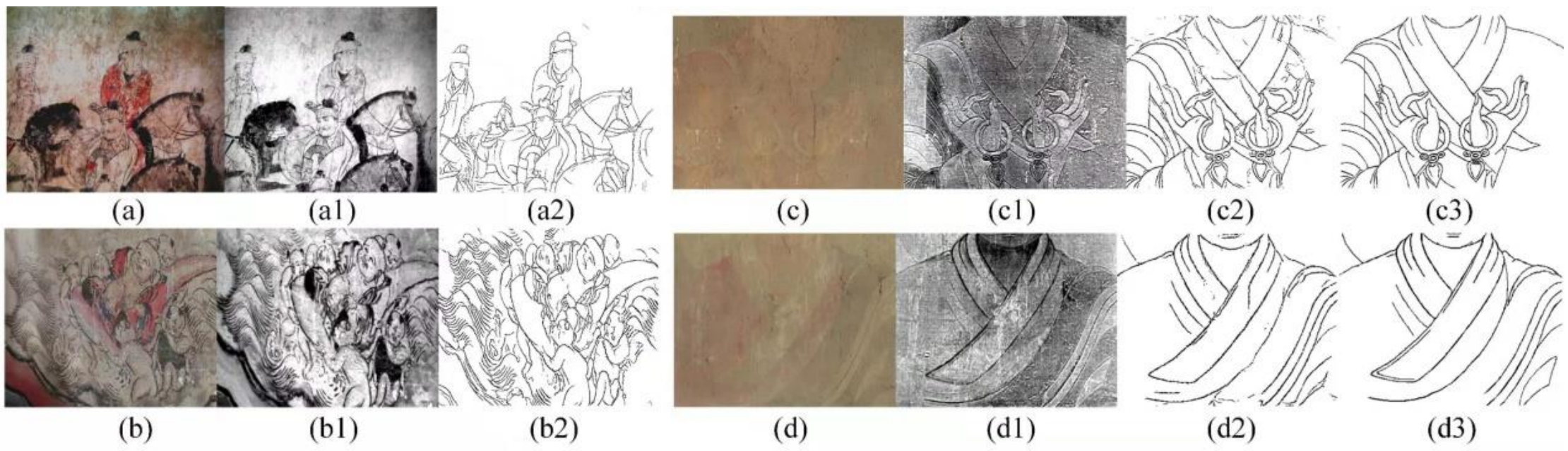

Figure 8: (a-d) Visible light images; (a1-d1) Manually selected hyperspectral band; (a2-d2) Sketch extracted by our method; (c3-d3) Ground truth.

The extraction results are shown in Fig. 8. For some unclear painted cultural relics in visible light, our method in combination with hyperspectral imaging technology could still extract clear sketches. Furthermore, the proposed meth-od could capture rich details, such as the costume details of the human in Fig. 8(a), the facial features in Fig. 8(b), the bracelet of Arhat in Fig. 8(c), and the stripes of clothes in Fig. 8(d). Furthermore, the noise suppression effect was very good for images with such serious diseases, such as the large number of wear tracks in Fig. 8(a), the disease in Fig. 8(b), and the corrosion marks and cracks in Fig. 8(c) and Fig. 8(d). This shows the generalization ability of the proposed method for the sketch extraction of hyperspectral data. The proposed method could still be extended to hyperspectral data sources to extract more accurate sketches for painted cultural relics images when the visible light is unclear. This has important significance for the complexity of cultural relics data and also provides higher practicality for actual cultural relics restoration and research work.

## Discussion and Conclusion

In this paper, we propose a novel hierarchical relic sketch extraction framework that consists of a detail-aware BDCN and MSU-Net. The proposed detail-aware BDCN integrates the FDoG and BDCN algorithms for the coarse sketch extraction. With the transfer learning technique, the detail-aware BDCN not only solves the limited training dataset problem in the relic sketch extraction but also preserves details and outlines well in the coarse sketch. The coarse sketch is used as the input of the proposed MSU-Net for refinement. Based on U-Net, the proposed MSU-Net fuses multiscale predictions from the decoder path, which not only utilizes feature maps from multiscale receptive fields in the encoder path for suppressing the effect of disease in the relic image but also utilizes multiscale feature maps in the decoder path for de-blurring and refining the sketch. The effectiveness of the proposed hierarchical sketch extraction framework was verified through experiments.

Moreover, we investigated the effects of two modules on the performance of the proposed method: detail-aware BDCN and MSU-Net. Experimental results showed that the detail-aware BDCN captures more detail without too many effects of noise and disease, which is an effective coarse sketch method. Compared with typical U-Net, MSU-Net shows a better performance in the noise suppression and line refinement. Our framework which combines with detail-aware BDCN and MSU-Net is the promising sketch extraction method.

Furthermore, we evaluated the generalization ability of the proposed method on hyperspectral relic images. Experimental results showed that the developed method is generalizable on the different source images, which provides more possibilities for sketch extraction of cultural relics images with complex conditions. In the future, we will continue to study the use of more effective bands of hyperspectral images to improve the accuracy of sketch extraction.

In summary, the proposed relic sketch extraction method achieved the best results among the compared state-of-the-art methods in each case and it can deal with the images with complex backgrounds, thereby indicating that the proposed method is a promising and effective relic sketch extraction method.

**Acknowledgments**

This work was supported by the National Key Research and Development Program of China (2017YFB1402103), the Key Research and Development Program of Shaanxi (2018ZDXM-GY-186), the Program for Changjiang Scholars and Innovative Research Team in University (IRT_17R87), the National Natural Science Foundation of China (62006-188), the Xi'an Key Laboratory of Intelligent Perception and Cultural Inheritance (2019219614SYS011CG033), Natural Science Basic Research Plan in Shaanxi Province of China (2019JQ-454, 2019JQ-294), China Postdoctoral Science Foundation (2018M643718).